\documentclass{mva_style}
\usepackage{graphicx}
\usepackage{color}
\usepackage{amsmath}
\usepackage{cite}
\usepackage{amsmath,amssymb,amsfonts}
\usepackage{algorithmic}
\usepackage{graphicx}
\usepackage{textcomp}
\usepackage{booktabs} 
\usepackage{array}
\usepackage{multirow}
\usepackage{colortbl}
\usepackage{textcomp}
\usepackage{color}
\usepackage[hidelinks,bookmarksnumbered=true,bookmarksopen=true]{hyperref}

\finalcopy %
\begin{document}
\title{Capturing Fine-Grained Alignments Improves\\3D Affordance Detection}

\author{
  Junsei Tokumitsu\textsuperscript{*}\\
  Keio AI Research Center\\
  {\tt toku.junsei@keio.jp}\\
  \and
  Yuiga Wada\textsuperscript{*}\\
  Keio AI Research Center\\
  {\tt yuiga@keio.jp}\\
}

\maketitle

\section*{\centering Abstract}
\textit{
    In this work, we address the challenge of affordance detection in 3D point clouds, a task that requires effectively capturing fine-grained alignments between point clouds and text. Existing methods often struggle to model such alignments, resulting in limited performance on standard benchmarks. A key limitation of these approaches is their reliance on simple cosine similarity between point cloud and text embeddings, which lacks the expressiveness needed for fine-grained reasoning.
    To address this limitation, we propose LM-AD, a novel method for affordance detection in 3D point clouds. 
    Moreover, we introduce the Affordance Query Module (AQM), which efficiently captures fine-grained alignment between point clouds and text by leveraging a pretrained language model.
    We demonstrated that our method outperformed existing approaches in terms of accuracy and mean Intersection over Union on the 3D AffordanceNet dataset.
}

\section{Introduction}
\vspace{-2mm}
Understanding affordances in object manipulation is essential for a wide range of robotic applications such as object grasp understanding \cite{a4t}, object recognition \cite{Thermos2017CVPR,Hou2021CVPR}, and agent activity recognition \cite{predictffromstatic,qi2017predicting,chen2023affordance}.
For safe robotic manipulation, grasp generation must take object affordances into account, especially when handling potentially hazardous objects, as failing to recognize appropriate affordances can pose serious safety risks to end users \cite{survey1, survey2, survey3}. 
For example, in a scenario where a robot is instructed to retrieve a pair of scissors, misinterpreting the affordance could result in the robot handing them over with the blades facing the user, resulting in a significant safety hazard.
\def\thefootnote{*}\footnotetext{Equal contribution}

Affordance detection methods can be broadly categorized into image-based and point-cloud-based approaches. Image-based methods predict affordances at the pixel level from an RGB image and a target affordance text.
In contrast, point-cloud-based methods aim to directly identify the corresponding points in a 3D point cloud based on the target affordance. 
Although image-based methods have been extensively explored, they pose a significant limitation in practical robotic applications: they require an additional transformation step to be used in downstream robotics tasks, as 2D predictions must be converted into 3D representations using depth information \cite{3daffordancenet, openad}. 
Since this conversion cannot always be performed with high quality, the additional step may lead to suboptimal performance in real-world scenarios. 
Therefore, point-cloud-based affordance detection plays a crucial role in practical robotics applications.

Point-cloud-based affordance detection is challenging because it requires efficiently capturing fine-grained alignments between point clouds and text, unlike typical 3D classification tasks \cite{ModelNet40,ScanObjectNN}, which solely focus on aligning entire point clouds with text. 
In fact, existing methods \cite{openad,openad-kd,TZSLPC,3DGenZ,ZSLPC} often fail to adequately capture the fine-grained alignment, resulting in accuracy below 50\% on standard benchmarks \cite{3daffordancenet}.
This performance bottleneck is partially because most of these methods (e.g. \cite{openad}, \cite{openad-kd}) rely only on cosine similarity between point cloud and text embeddings, which lacks the expressiveness needed to model fine-grained alignment.

\begin{figure}[t]
    \centering
    \includegraphics[width=\linewidth]{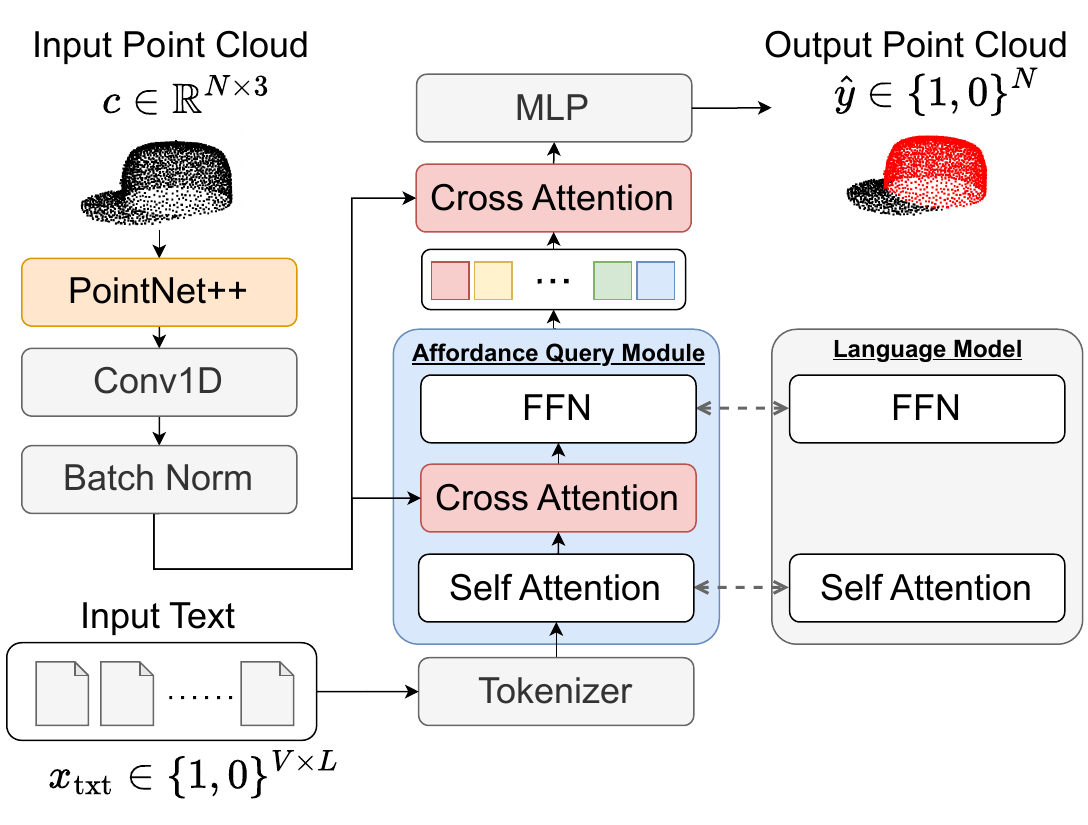}
    \large
    \caption{Overview of the proposed LM-AD and Affordance Query Module (AQM).}
    \label{fig:model}
    \vspace{-6mm}
\end{figure}
To address this limitation, we propose LM-AD, a novel method for affordance detection in 3D point clouds. 
Moreover, we introduce the Affordance Query Module (AQM), which efficiently captures fine-grained alignment between point clouds and text by leveraging a pretrained language model (LM).
Our approach is inspired by recent advances in vision-language models such as BLIP-2 \cite{blip2} and Flamingo \cite{flamingo}, which are designed to efficiently learn the alignment between two modalities.
Previous studies \cite{blip2, flamingo} have demonstrated that inserting trainable cross-attention layers \cite{transformer} into a pretrained LM \cite{bert, chinchilla} enables fine-grained interaction between visual and textual features.
Building on this insight, our method repeatedly fuses LM features with point cloud features to achieve more effective alignment, unlike existing affordance detection methods that rely solely on cosine similarity to capture relationships between the two modalities \cite{openad, openad-kd}.
Our key contributions are as follows:
\vspace{-2mm}
\begin{enumerate}
    \setlength{\parskip}{0.5mm} %
    \setlength{\itemsep}{0.2mm} %
\item We propose LM-AD, a novel method for affordance detection in 3D point clouds. 
\item We introduce the Affordance Query Module (AQM), which efficiently captures fine-grained alignment between point clouds and text by leveraging a pretrained language model (LM).
\item We demonstrated that our method outperformed existing approaches in terms of accuracy and mean Intersection over Union (mIoU) on the 3D AffordanceNet dataset.
\end{enumerate}
\vspace{-4mm}

\section{Related Work}
\vspace{-1.5mm}

\vspace{-1.5mm}
\subsection{Image-based Affordance Detection}
\vspace{-1.5mm}
Image-based methods predict affordances at the pixel level from an RGB image and a target affordance text \cite{7759429,do2018affordancenet,8206484,roy2016multi,thermos2020deep,chen2022cerberus,luo2022learning,qian2024affordancellm}.
Various studies have proposed affordance detection models based on either CNNs or Transformers. 
For instance, Nguyen et al \cite{7759429} proposed a simple CNN architecture for affordance detection, whereas AffordanceNet \cite{do2018affordancenet} employs a two-branch network: one branch performs object detection, while the other assigns the most appropriate affordance label to each pixel within the detected object.
In \cite{8206484}, the author proposed a method incorporating Conditional Random Fields into a CNN for improved affordance detection. 
In contrast, Roy et al. \cite{roy2016multi} generate depth and surface normal maps from RGB images and use these additional modalities to enhance detection performance.

In addition to CNN-based methods, Transformer-based approaches have also been actively explored. A notable example is Cerberus Transformer \cite{chen2022cerberus}, which leverages a Transformer architecture to jointly detect affordance labels, object attributes, and semantic categories from a single image.
On the other hand, AffordanceLLM \cite{qian2024affordancellm} utilizes Multimodal Large Language Models (MLLMs), such as LLaVA \cite{llava}, to predict affordances from visual and textual inputs.
Although image-based methods have shown compelling results, they present a significant limitation in practical robotic applications. As pointed out in \cite{openad}, these methods require an additional transformation step to convert 2D predictions into 3D representations.

\vspace{-1.5mm}
\subsection{Point-Cloud-based Affordance Detection}
\vspace{-1.5mm}
Point-cloud-based methods aim to directly identify the corresponding points in a 3D point cloud based on the target affordance. 
Classic approaches employ techniques such as $k$-means clustering and logistic regression. For instance, Kim et al. \cite{kim2014semantic} proposed a method that classifies affordances using logistic regression, whereas Kokic et al. \cite{kokic2017affordance} addressed task-specific grasping by leveraging CNNs to model the relationships among tasks, objects, and grasp actions.

More recent methods utilize point cloud encoders \cite{pointnet, pointnet-pp, dgcnn, point-transformer} and text encoders \cite{bert, roberta, deberta} to handle cross-modal embeddings. 
OpenAD \cite{openad} performs point-cloud-wise affordance detection by computing the correlation matrix, in which each element is just the cosine similarity between point cloud embeddings extracted by PointNet++ \cite{pointnet-pp} and text embeddings extracted by CLIP \cite{radford_clip_2021}.
In contrast, OpenAD-KD \cite{openad-kd} extends OpenAD by employing knowledge distillation (KD) to address its limitations in handling out-of-domain 3D objects and previously unseen affordances.
As aforementioned in Section 1, these methods often struggle to capture fine-grained alignment, resulting in accuracy below 50\% on standard benchmarks (e.g. \cite{3daffordancenet}).
This is partly due to their reliance on cosine similarity (e.g., \cite{openad, openad-kd}), which may be suboptimal for this task.

\vspace{-1.5mm}
\subsection{Multimodal Alignment and Fusion}
\vspace{-1.5mm}
Li et al. \cite{survey-fuse} introduced a novel taxonomy that categorizes cross-modal fusion strategies into three types: \textit{two-tower}, \textit{two-leg}, and \textit{one-tower} models.
First, \textit{two-tower} models \cite{radford_clip_2021,jia2021ALIGN,liang2024multimodal,vasilakis2024instrument,xu2023bridgetower,su2023beyond,du2023touchformer,chen2024mixtower,fei2022towards,wen2024multimodal,tu2022crossmodal,yuan2021medication} process two modalities separately and combine their embeddings using simple operations. Second, \textit{two-leg} models \cite{Allaire2012FusingIF, Badrinarayanan2015SegNetAD,danapal2020sensofusion,Guo2023AMF,Jaiswal2015LearningTC,wada2024,Li2020HierarchicalFF,Li2018DenseFuseAF,Mai2019DivideCA,Makris2011AHF,Missaoui2010ModelLF,Rvid2019TowardsRS,Steinbaeck2018DesignOA,Uezato2020GuidedDD,Wei2021DecisionLevelDF,kim_vilt_2021} also extract embeddings from each modality independently but use additional neural networks to perform fusion. 
A notable limitation of these approaches is that such fusion operations often fail to achieve deep and effective fusion due to insufficient interaction between the modalities.

\textit{One-tower} models \cite{bao_vlmo_nodate,li_blip_2022,Li2023BLIP2,chen2023instructblip,zhu2023minigpt4enhancingvisionlanguageunderstanding,wang2022simvlm,wang2024qwen2vlenhancingvisionlanguagemodels,bai2023qwenvl,flamingo} use a unified architecture to jointly embed two input modalities, often leveraging a pretrained language model (LM). A representative example is BLIP-2 \cite{blip2}, which bridges the modality gap with a lightweight Querying Transformer, which uses a set of learnable query vectors to extract visual features.
Moreover, Alayrac et al. \cite{flamingo} demonstrated that inserting simple trainable layers \cite{transformer} into a pretrained LM \cite{bert, chinchilla} enables learning the fine-grained alignment between visual and textual features.

\vspace{-3mm}
\section{Problem Settings}
\vspace{-3mm}
We address the task of affordance detection for 3D point clouds \cite{openad, openad-kd}. In this task, it is desirable to predict the appropriate affordances for the given 3D point cloud objects. 

The inputs include a 3D point cloud $ c \in \mathbb{R}^{N \times 3} $ comprised of $ N $ points and a text related to affordances $ x_{\text{txt}} \in \{1, 0\}^{V \times L} $. The output is points $ \hat{y} \in \{1, 0\}^{N} $ corresponding to $ x_{\text{txt}} $. Here, $ V $ represents the vocabulary size, and $ L $ denotes the maximum token length.
Following existing studies \cite{openad, openad-kd}, we assumed that $x_{\text{txt}}$ is composed of a single word, rather than sentences.

\vspace{-1.5mm}
\section{Methodology}
\vspace{-1.5mm}
We propose LM-AD (Language Model-guided Affordance Detection), a novel method for affordance detection in 3D point clouds.  Fig. \ref{fig:model} shows the overview of the proposed method.
Our method is inspired by existing vision-language models such as BLIP-2 \cite{blip2} and Flamingo \cite{flamingo}, which are designed to efficiently learn the alignment between two modalities. Moreover, to achieve the fine-grained alignment between point clouds and texts, we introduce the Affordance Query Module (AQM), which efficiently captures such alignment by leveraging a pretrained language model (LM).
Unlike existing methods that rely solely on cosine similarity to capture relationships between the two modalities \cite{openad, openad-kd}, our method repeatedly fuses LM features with point cloud features to achieve more effective alignment.

\vspace{-1.5mm}
\subsection{Point Cloud Feature Extraction}
\vspace{-1.5mm}
First, using PointNet++ \cite{pointnet-pp}, we extract an embedding $ {h}_{p} \in \mathbb{R}^{N \times d_{P}} $ from $ {c} $, where $ d_{P} $ denotes the output dimension of PointNet++. 
We adopt PointNet++ as the point cloud encoder, because it is standard in this field \cite{openad, openad-kd, TZSLPC, 3DGenZ}.
Subsequently, to enhance the capture of relationships between nodes, we feed  $ h_{p} $  into a one-dimensional convolution followed by batch normalization, resulting in the final embedding $h_{c} \in \mathbb{R}^{N \times d_{P}} $.

\vspace{-1.5mm}
\subsection{Affordance Query Module}
\vspace{-1.5mm}

We introduce the Affordance Query Module (AQM), which efficiently captures fine-grained alignment between point clouds and text by leveraging a pretrained language model (LM).
Unlike 3D classification tasks \cite{ModelNet40,ScanObjectNN}, which focus on aligning entire point clouds with text, this task requires efficiently capturing fine-grained alignments.
However, most existing methods \cite{openad, openad-kd} rely solely on computing the cosine similarity between point clouds and text embeddings, which may be suboptimal for capturing such alignments. 
To address this limitation, we take inspiration from recent advances in vision-language models. Previous works \cite{blip2, flamingo} showed that inserting trainable cross-attention layers \cite{transformer} into a pretrained LM \cite{bert, chinchilla} can enable fine-grained interaction between visual and textual features.

Building on this insight, AQM leverages a pretrained LM by inserting cross-attention layers into its architecture. Specifically, AQM is built upon an LM consisting of $L_{\text{LM}}$ layers, and comprises $L_{\text{LM}}$ corresponding AQM blocks. Each block retains the original LM structure, but with a cross-attention layer inserted between the self-attention and feed-forward network layers.

The AQM takes the text $ x_{\text{txt}} $ and point cloud features $ h_c $ as input, and outputs a representation $ g $ that captures fine-grained alignment between the two modalities. The $ i $-th AQM block first feeds its input $ x^{(i)} $ into the self-attention mechanism of the pretrained LM to obtain an intermediate feature $ h^{(i)} $ as follows:  
\begin{align}
h^{(i)} = \text{SelfAttn}_{\text{LM}}(x^{(i)}),
\end{align}  
where the initial input is defined as $ x^{(1)} = x_{\text{txt}} $.  
Subsequently, we compute cross-modal features $ g^{(i)} $ by applying a cross-attention mechanism followed by the feed-forward network:  
\begin{align}
g^{(i)} = \text{FFN}_{\text{LM}}\left(\text{CrossAttn}\left(h^{(i)},h_c\right)\right),
\end{align}  
Here, the query is given by $h^{(i)}$, whereas both the key and value are derived from $ h_c $.
The final output of AQM, denoted as $g$, is defined as the output of the last AQM block: $g = g^{(L_{\text{LM}})}$.
Note that we employed BERT~\cite{bert} as the pretrained language model, following prior work (e.g.,~\cite{blip2}).
We used the base configuration with 12 layers ($L_\mathrm{LM} = 12$), a hidden size of 768, and the standard BERT tokenizer.

\vspace{-1.5mm}
\subsection{Affordance Prediction}
\vspace{-1.5mm}
To decode the richly aligned features provided by AQM, we feed the output of AQM $ g $ and the point cloud features  $ h_c $ into a cross-attention mechanism followed by an MLP.  
Specifically, the final output $ \hat{y} $ is computed as follows:  
\begin{align}
\hat{y} = \text{MLP}\bigl(\text{CrossAttn}\left(h_c,\: g\right)\bigr).
\end{align}  
Here, the query is $ h_c $, while both the key and value are $ g $. 
For the loss function, we adopt cross-entropy, which is a standard choice for this task.

\vspace{-1.5mm}
\section{Experiments}
\vspace{-1.5mm}

\subsection{Experimental Setup}
\vspace{-1.5mm}

Following prior work \cite{openad, openad-kd}, we used the 3D AffordanceNet dataset \cite{3daffordancenet} to evaluate the proposed method.
The dataset includes two tasks: full-shape and partial-view. In the full-shape task, the model is expected to predict affordances from complete point clouds. In contrast, the partial-view task requires the model to predict based on incomplete point clouds. This setup is particularly important in robotics, as it reflects the limited observation capabilities of real-world robotic systems.
The 3D AffordanceNet contains 23 object categories of 3D point cloud objects, totaling 22,949 objects and 18 affordance labels. 

We employ TZSLPC\cite{TZSLPC}, 3DGenZ\cite{3DGenZ}, ZSLPC\cite{ZSLPC}, OpenAD\cite{openad}, and OpenAD-KD\cite{openad-kd} as baseline methods because they are representative methods for 3D affordance detection.
For evaluating the method, we use the standard metrics for this task, which include the mean Intersection over Union (mIoU), overall accuracy overall points (Acc), and mean accuracy over all classes (mAcc).

We split the dataset into train, validation, and test sets with a ratio of 70\%, 10\%, and 20\%, respectively, according to the shape semantic category \cite{openad,openad-kd,3daffordancenet}.
We used the training set to train the model, the validation set for hyperparameter tuning, and the test set to evaluate the performance of the model.
We trained our proposed method on an NVIDIA RTX A6000 (with 48GB of memory) and an Intel Core i7-14700 (with 64GB of memory).
The training time for the proposed method was approximately 11.6 hours, and the inference time per sample was approximately 72ms.

\vspace{-1.5mm}
\subsection{Quantitative and Qualitative Results}
\vspace{-1.5mm}
\begin{table}[t]
\caption{Quantitative comparison with baselines.}
\vspace{-6mm}
\begin{center}
\resizebox{\columnwidth}{!}{%
\begin{tabular}{llccc}
\toprule
Task & Method  &  mIoU & Acc & mAcc \\
\midrule
\multirow{6}{*}{\rotatebox{90}{Full-shape}} 
& TZSLPC\cite{TZSLPC} & 3.86 & 42.97 & 10.37 \\
& 3DGenZ~\cite{3DGenZ} & 6.46 & 45.47 & 18.33 \\
& ZSLPC~\cite{ZSLPC} & 9.97 & 40.13 & 18.70 \\
& OpenAD~\cite{openad} & 14.37 & 46.31 & 19.51 \\
& OpenAD-KD~\cite{openad-kd} & {22.33} & {49.72} & {34.29} \\
& \bf{LM-AD~\small{(Ours)}} & \bf{41.98} & \bf{68.60} & \bf{68.89} \\
\midrule
\multirow{6}{*}{\rotatebox{90}{Partial-view}} 
& TZSLPC~\cite{TZSLPC} & 4.14 & 42.76 & 8.49 \\
& 3DGenZ~\cite{3DGenZ} & 6.03 & 45.24 & 15.86 \\
& ZSLPC~\cite{ZSLPC} & 9.52 & 40.91 & 17.16 \\
& OpenAD~\cite{openad}  & 12.50 & 45.25 & 17.37 \\
& OpenAD-KD~\cite{openad-kd}  &{20.48} & {48.72} & {32.86} \\
& \bf{LM-AD~\small{(Ours)}}  &\textbf{35.26} & \textbf{62.65} & \textbf{59.09} \\
\bottomrule
\end{tabular}
}
\end{center}
\label{tab:results}
\vspace{-7mm}
\end{table}

\begin{table}[t]
\vspace{-1.5mm}
\caption{Ablation study.}
\vspace{-1.5mm}
\begin{center}
\begin{tabular}{ccccc}
\toprule
Method & AQM & mIoU & Acc & mAcc \\
\midrule
(i) & & 19.38 & 48.14 & 27.31 \\
(ii) & $\checkmark$ & \bf{41.98} & \bf{68.60} & \bf{68.89} \\
\bottomrule
\end{tabular}
\end{center}
\label{tab:ablation}
\vspace{-8mm}
\end{table}

Table \ref{tab:results} presents the quantitative comparison results between the baselines and our proposed method on the 3D AffordanceNet. 
Our proposed method achieved mIoU scores of 41.98 and 35.26, Acc scores of 68.60 and 62.65, and mAcc scores of 68.89 and 59.09 on the full-shape and partial-view tasks, respectively. 
These results demonstrated that our method outperformed OpenAD by margins of 27.62, 21.90, and 49.38 points in mIoU, Acc, and mAcc on the full-shape task, and by 22.76, 17.40, and 41.72 points, respectively, on the partial-view task.
Similarly, compared to OpenAD-KD, our method also showed improvements of 19.65, 18.88, and 34.60 points for the full-shape task, and 14.87, 13.93, and 26.23 points for the partial-view task, in terms of mIoU, Acc, and mAcc, respectively.

\begin{figure}[t]
    \centering
    \includegraphics[width=\linewidth]{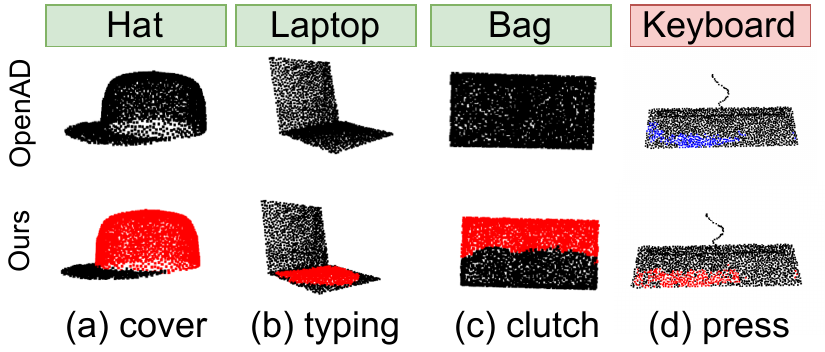}
    \large
    \caption{Qualitative results. Cases (a), (b), and (c) show successful examples, while Case (d) shows a failure example.}
    \label{fig:results}
    \vspace{-6mm}
\end{figure}

Fig.~\ref{fig:results} shows qualitative results of our proposed method. Cases (a), (b), and (c) illustrate successful examples, while Case (d) presents a failure case.
In Case (a), our method successfully detected the ``cover'' affordance from a hat. This was particularly challenging because it was a zero-shot setting. While OpenAD failed to predict the affordance correctly in this instance, our method produced an appropriate prediction.
Case (b) shows the detection results for the ``typing'' affordance from a laptop.
Similar to Case (a), our method succeeded in identifying the appropriate region, whereas OpenAD failed.
In Case (c), our method appropriately predicted the ``clutch'' affordance from a bag, while OpenAD produced an inaccurate prediction.

In contrast, Case (d) illustrates a failure case involving a keyboard, where the objective was to detect the ``press'' affordance. In this case, the model was expected to identify all of the keys; however, both our method and the baseline failed.
As suggested by \cite{openad-kd}, this failure is likely due to the visual similarity between the keyboard and the bag, as both share rectangular shapes. Such resemblance may hinder the appropriate prediction of corresponding affordances when relying solely on point cloud geometry.

\vspace{-2mm}
\subsection{Ablation Study}
\vspace{-2mm}

We conducted ablation studies to evaluate the impact of the AQM by replacing it with simple cross-attention layers while keeping the number of layers unchanged. Table~\ref{tab:ablation} presents the results.
Method (i), which uses the simplified cross-attention to fuse language and point cloud features, achieved 19.38 mIoU, 48.14 Acc, and 27.31 mAcc. These results represent significant drops of 22.60, 20.46, and 41.58 points compared to Method (ii), which corresponds to the original LM-AD.
These results indicate that our AQM contributed significantly to the overall performance.

\vspace{-2mm}
\section{Conclusion}
\vspace{-2mm}

In this study, we addressed the task of affordance detection for 3D point clouds. The contributions of this paper are as follows: (i) We proposed LM-AD, a novel method for affordance detection in 3D point clouds. (ii) We introduced the Affordance Query Module, which efficiently captures fine-grained alignment between point clouds and text by leveraging a pretrained LM. (iii) We demonstrated that our method outperformed existing approaches in terms of accuracy and mIoU on the 3D AffordanceNet dataset.

\vspace{-2mm}
\section*{Acknowledgements}
\vspace{-2mm}
\noindent The authors would like to express their sincere gratitude to Professor Komei Sugiura and Professor Hideo Saito of the Keio AI Research Center for their continuous guidance and support throughout this research.

\bibliographystyle{ieeetr}
\bibliography{ref}

\end{document}